\documentclass{article}
\usepackage{spconf,amsmath,graphicx}

\usepackage{caption}
\usepackage{wrapfig}
\usepackage{setspace}
\usepackage{cleveref}
\usepackage{enumitem}

\usepackage{multirow}
\usepackage{multicol}
\usepackage{xcolor}

\makeatletter
\renewcommand{\section}{\@startsection
  {section}%
  {1}%
  {}%
  {-0.5\baselineskip}%
  {0.2\baselineskip}%
  {}}%

\renewcommand{\subsection}{\@startsection
  {subsection}%
  {2}%
  {}%
  {-0.1\baselineskip}%
  {0.1\baselineskip}%
  {}}%

\renewcommand{\subsubsection}{\@startsection
  {subsubsection}%
  {3}%
  {}%
  {-0.2\baselineskip}%
  {0.2\baselineskip}%
  {}}%

\g@addto@macro\normalsize{%
  \setlength\abovedisplayskip{5pt plus 2pt minus 2pt}
  \setlength\belowdisplayskip{5pt plus 2pt minus 2pt}
  \setlength\abovedisplayshortskip{4pt plus 2pt minus 2pt}
  \setlength\belowdisplayshortskip{4pt plus 2pt minus 2pt}
}

\makeatother

\Crefname{equation}{Eq.}{Eqs.}
\Crefname{figure}{Fig.}{Figs.}
\Crefname{tabular}{Tab.}{Tabs.}

\DeclareMathOperator*{\argmax}{arg\,max\hspace{2mm}}

\newcommand\numberthis{\addtocounter{equation}{1}\tag{\theequation}}

\def\L{{\cal L}}
\def\B{{\cal B}}

\title{On Language Model Integration for RNN Transducer based\\ Speech Recognition}
\name{Wei Zhou$^{1,2}$, Zuoyun Zheng$^{1}$, Ralf Schl\"uter$^{1,2}$, Hermann Ney$^{1,2}$}
\address{
$^1$Human Language Technology and Pattern Recognition, Computer Science Department,\\
  RWTH Aachen University, 52074 Aachen, Germany \\
$^2$AppTek GmbH, 52062 Aachen, Germany}

\begin{document}
%
\maketitle
\begin{abstract}
The mismatch between an external language model (LM) and the implicitly learned internal LM (ILM) of RNN-Transducer (RNN-T) can limit the performance of LM integration such as simple shallow fusion.
A Bayesian interpretation suggests to remove this sequence prior as ILM correction.
In this work, we study various ILM correction-based LM integration methods formulated in a common RNN-T framework. 
We provide a decoding interpretation on two major reasons for performance improvement with ILM correction, which is further experimentally verified with detailed analysis.
We also propose an exact-ILM training framework by extending the proof given in the hybrid autoregressive transducer, which enables a theoretical justification for other ILM approaches.
Systematic comparison is conducted for both in-domain and cross-domain evaluation on the Librispeech and TED-LIUM Release 2 corpora, respectively. Our proposed exact-ILM training can further improve the best ILM method.

\end{abstract}
\section{Introduction \& Related Work}
End-to-end (E2E) speech recognition has shown great simplicity and state-of-the-art performance \cite{zoltan2020swb, Gulati20conformer}. 
Common E2E approaches include connectionist temporal classification (CTC) \cite{graves2016ctc}, recurrent neural network transducer (RNN-T) \cite{graves2012rnnt} and attention-based encoder-decoder models \cite{bahdanau2016end, chan2016listen}.
While E2E models are only trained on paired audio-transcriptions in general, an external language model (LM) trained on much larger amount of text data (from possibly better-matched domain) can further boost the performance. 
Without modifying the model structure, shallow fusion (SF) \cite{gulcehre2015shallowFusion} has been a widely-used effective LM integration approach for E2E models, which simply applies a log-linear model combination. 

However, with context dependency directly included in the posterior distribution, both RNN-T and attention models implicitly learn an internal LM (ILM) as a sequence prior restricted to the audio transcription only. This ILM usually has a strong mismatch to the external LM, which can limit the performance of LM integration such as simple SF. Existing approaches to handle the ILM fall into 3 major categories:
\vspace{-1.2mm}
\begin{itemize}[leftmargin=*, itemsep=-0.5mm]
\item ILM suppression: The ILM can be suppressed during the E2E model training via limiting context/model size \cite{zeineldeen2021ilm} or introducing an external LM at early stage \cite{michel20lm_integration}.
\item ILM correction: The ILM can be estimated in various ways \cite{McDermott2019DR, variani2020hat, ILME, zeyer2021libspTransducer, zeineldeen2021ilm} and then corrected from the posterior in decoding, which fits into a Bayesian interpretation.
\item ILM adaptation: The ILM can be adapted on the same text data used by the external LM to alleviate the mismatch. This can be done via text to speech \cite{rnntTTS,rnntTTSmap} to train the E2E model or partial model update directly on the text \cite{rnntTextAdapt2021}.
\end{itemize}
\vspace{-1.2mm}
ILM suppression requires training modification and performs similarly as ILM correction \cite{zeineldeen2021ilm}. 
ILM adaptation has a even higher complexity and may still need ILM correction for further improvement \cite{rnntTTS}. Therefore, ILM correction appears to be the most simple and effective approach for LM integration, which also has a better mathematical justification \cite{McDermott2019DR, variani2020hat}.

The density ratio \cite{McDermott2019DR} approach estimates the ILM directly from the transcription statistics. The hybrid autoregressive transducer (HAT) \cite{variani2020hat} proposed to estimate the ILM from the transducer neural network (NN) by excluding the impact of the encoder, which is justified with a detailed proof under certain approximation. This approach is further investigated in \cite{ILME} and extended by including the average encoder output in \cite{zeyer2021libspTransducer}. 
In \cite{variani2020hat, Meng2021ILMT}, ILM training (ILMT) is applied to include the ILM into the E2E model training for joint optimization.
Recently, \cite{zeineldeen2021ilm} conducted a comprehensive study on ILM methods for the attention model, and proposed the mini-LSTM approach that outperforms all previous methods.

In this work, we study various ILM correction-based LM integration methods for the RNN-T model.
We formulate different ILM approaches proposed for different E2E systems into a common RNN-T framework. 
We provide a decoding interpretation on two major reasons for performance improvement with ILM correction, which is further experimentally verified with detailed analysis.
Additionally, we extend the HAT proof \cite{variani2020hat} and propose an exact-ILM training framework which enables a theoretical justification for other ILM approaches.
Systematic comparison is conducted for both in-domain and cross-domain evaluation on the Librispeech \cite{libsp} and TED-LIUM Release 2 (TLv2) \cite{tedlium2} corpora, respectively. The effect of ILMT on these approaches is also investigated.

\section{RNN-Transducer}
Given a speech utterance, let $a_1^S$ denote the output (sub)word sequence of length $S$ from a vocabulary $V$. Let $X$ denote the corresponding acoustic feature sequence and $h_1^T = f^{\text{enc}}(X)$ denote the encoder output, which transforms the input into high-level representations.
The RNN-T model \cite{graves2012rnnt} defines the sequence posterior as:\\
\scalebox{0.9}{\parbox{1.11\linewidth}{%
\begin{align*}
P_{\text{RNNT}}(a_1^S|X) &= \sum_{y_1^{U=T+S}: \B^{-1}(a_1^S)} P_{\text{RNNT}}(y_1^U | h_1^T) \\
&= \sum_{y_1^U:\B^{-1}(a_1^S)} \prod_{u=1}^{U=T+S} P_{\text{RNNT}}(y_u | \B(y_1^{u-1}), h_1^T) \numberthis \label{eq:rnnt}
\end{align*}}}
Here $y_1^U$ is the blank $\epsilon$-augmented alignment sequence, which is uniquely mapped to $a_1^S$ via the collapsing function $\B$ to remove all blanks.
The common NN structure of RNN-T contains an encoder $f^{\text{enc}}$, a prediction network $f^{\text{pred}}$ and a joint network $J$ followed by a softmax activation. We denote $\theta_{\text{RNNT}}$ as the RNN-T NN parameters.
The probability in \Cref{eq:rnnt} can also be represented from the lattice representation of the RNN-T topology \cite{graves2012rnnt}. 
By denoting $y_1^{u-1}$ as a path reaching a node $(t, s-1)$, we have:\\
\scalebox{0.9}{\parbox{1.11\linewidth}{%
\begin{align*}
P_{\text{RNNT}}(y_u | \B(y_1^{u-1}), h_1^T) &= P_{\text{RNNT}}(y_u | a_1^{s-1}, h_t) \numberthis \label{eq:rnntNode} \\
&= \text{Softmax} \left[ J\left( f^{\text{pred}}(a_1^{s-1}), f_t^{\text{enc}}(X) \right) \right] 
\end{align*}}}
where $y_1^u$ reaches $(t+1, s-1)$ if $y_u=\epsilon$ or $(t, s)$ otherwise.

Without an external LM, $P_{\text{RNNT}}(a_1^S|X)$ can be directly applied into the maximum a posteriori (MAP) decoding to obtain the optimal output sequence:\\
\scalebox{0.9}{\parbox{1.11\linewidth}{%
\begin{align*}
X \rightarrow \tilde{a}_1^{\tilde{S}} = \argmax_{a_1^S, S} P(a_1^S|X) \numberthis \label{eq:MAP}
\end{align*}}}
Within a Bayesian framework, to integrate the RNN-T model and an external LM jointly into \Cref{eq:MAP} as modularized components, the Bayes' theorem needs to be applied:\\
\scalebox{0.9}{\parbox{1.11\linewidth}{%
\begin{align*}
X \rightarrow \tilde{a}_1^{\tilde{S}} &= \argmax_{a_1^S, S} P(a_1^S) \cdot P(X | a_1^S) \numberthis \label{eq:MAP-joint}\\
&= \argmax_{a_1^S, S} P^{{\lambda}1}_{\text{LM}}(a_1^S) \cdot \frac{P_{\text{RNNT}}(a_1^S|X)}{P^{{\lambda}2}_{\text{RNNT-ILM}}(a_1^S)} \numberthis \label{eq:MAP-ILM}
\end{align*}}} 
which suggests the removal of the RNN-T model's internal sequence prior $P_{\text{RNNT-ILM}}$. 
Here ${\lambda}1$ and ${\lambda}2$ are scales applied in common practice.  The SF approach \cite{gulcehre2015shallowFusion} essentially omits $P_{\text{RNNT-ILM}}$ completely with ${\lambda}2=0$.

\section{Internal LM}
$P_{\text{RNNT-ILM}}$ is effectively the ILM, which is implicitly learned and contained in $P_{\text{RNNT}}$ after training. 
It should be most accurately obtained by marginalization $\sum_{X} P_{\text{RNNT}}(a_1^S|X)P(X)$.
Since the exact summation is intractable, an estimated $P_{\text{ILM}}$ is usually applied for approximation.

\subsection{ILM Estimation}
One straightforward way is to assume that $P_{\text{RNNT-ILM}}$ closely captures the statistics of the acoustic training transcription. This is the density ratio approach \cite{McDermott2019DR} which trains a separate $P_{\text{ILM}}$ on the audio transcription. 

To be more consistent with the $P_{\text{RNNT}}$ computation, another popular direction is to partially reuse the RNN-T NN for computing $P_{\text{ILM}}$. A general formulation can be given as:\\
\scalebox{0.9}{\parbox{1.11\linewidth}{%
\begin{align*}
P_{\text{ILM}}(a_1^S) &= \prod_{s=1}^S P_{\text{ILM}}(a_s | a_1^{s-1}) = \prod_{s=1}^S P'(a_s| a_1^{s-1}, h') \numberthis \label{eq:h-ILM}
\end{align*}}}
where $P'$ is defined over $V$ and $h'$ is some global representation. 
Here $P'$ shows a strong correspondence to \Cref{eq:rnntNode} except that $P_{\text{RNNT}}$ is defined over $V \cup \{\epsilon\}$ in general. In \cite{variani2020hat}, a separate blank distribution in $P_{\text{RNNT}}$  is proposed to directly use its label distribution for $P'$. However, the same can always be achieved by a simple renormalization:\\
\scalebox{0.9}{\parbox{1.11\linewidth}{%
\begin{align*}
P'(a_s| a_1^{s-1}, h') &= \frac{P_{\text{RNNT}}(a_s | a_1^{s-1}, h')}{ 1 - P_{\text{RNNT}}(\epsilon | a_1^{s-1}, h')} \numberthis \label{eq:renorm} \\
&= \text{Softmax}\left[ J_{\backslash\epsilon}\left( f^{\text{pred}}(a_1^{s-1}), h' \right) \right]
\end{align*}}}
where $J_{\backslash\epsilon}$ stands for the joint network $J$ excluding the blank logit output. This is also done in \cite{ILME}.
In fact, we recommend to always use a single distribution for $P_{\text{RNNT}}$, since it is equivalent to the separate distributions in \cite{variani2020hat} after renormalization, but has the advantage of further discrimination between blank and speech labels.
This can partially prevent blank being too dominant which may lead to a sensitive decoding behavior with high deletion errors as observed in \cite{mWER_HAT}.
Existing ILM estimation approaches are then categorized by the way of representing $h'$:
\vspace{-1.2mm}
\begin{enumerate}[leftmargin=*, itemsep=-0.5mm]
\item $h'_{\text{zero}}: h' = \vec{0}$ \hspace{2mm} \cite{variani2020hat, ILME} 
\item $h'_{\text{avg}}: h' = \text{mean}(h_1^T)$ \hspace{2mm} \cite{zeyer2021libspTransducer, zeineldeen2021ilm}
\item $h'_{a_1^{s-1}}: h' = f_{\theta_{\text{ILM}}}(a_1^{s-1})$
where an additional NN $f_{\theta_{\text{ILM}}}$ is introduced to generate $h'$ based on $a_1^{s-1}$. The mini-LSTM method \cite{zeineldeen2021ilm} falls into this category (denoted as $h'_{\text{mini-LSTM}}$).
\end{enumerate}
\vspace{-1.2mm}
All these approaches are based on fixed $\theta_{\text{RNNT}}$.
For the $h'_{a_1^{s-1}}$ approach, an LM-like loss $\L_{\text{ILM}} = -\log P_{\text{ILM}}(a_1^S)$ based on \Cref{eq:h-ILM} and \Cref{eq:renorm} is used to train the additional $f_{\theta_{\text{ILM}}}$ over the audio transcription. This effectively combines the advantage of using transcription statistics and partial RNN-T NN.

\subsection{ILM Training}
The RNN-T model is commonly trained with a full-sum loss $\L_{\text{RNNT}} = -\log P_{\text{RNNT}}(a_1^S|X)$ over all alignment paths as in \Cref{eq:rnnt}. 
When reusing the RNN-T NN for $P_{\text{ILM}}$, one can also combine $\L_{\text{RNNT}}$ and $\L_{\text{ILM}}$ as a multi-task training to train all parameters  including $\theta_{\text{RNNT}}$ jointly:\\
\scalebox{0.9}{\parbox{1.11\linewidth}{%
\begin{align*}
\L_{\text{ILMT}} = \L_{\text{RNNT}} + \alpha \L_{\text{ILM}}
\end{align*}}}
where $\alpha$ is a scaling factor. This joint training is applied to the $h'_{\text{zero}}$ approach in \cite{variani2020hat, Meng2021ILMT}, which we also apply to the $h'_{\text{avg}}$ and $h'_{a_1^{s-1}}$ approaches in this work.


\subsection{Decoding Interpretation}
\label{sec:decode}
Since both $P_{\text{LM}}$ and $P_{\text{ILM}}$ are only defined over $V$, we can further expand \Cref{eq:MAP-ILM} as:\\
\scalebox{0.9}{\parbox{1.11\linewidth}{%
\begin{align*}
\argmax_{a_1^S, S}& \hspace{-2mm} \sum_{y_1^U:\B^{-1}(a_1^S)} \prod_{u=1}^{U} P_{\text{RNNT}}(y_u | \B(y_1^{u-1}), h_1^T) \cdot Q(y_u| \B(y_1^{u-1})) \\
\text{with} \hspace{2mm} &Q(y_u| \B(y_1^{u-1})) =
\begin{cases}
1, &y_u = \epsilon \\
\frac{P^{{\lambda}1}_{\text{LM}}(y_u | \B(y_1^{u-1}))}{P^{{\lambda}2}_{\text{ILM}}(y_u | \B(y_1^{u-1}))}, &y_u \neq \epsilon
\end{cases} 
\end{align*}}}
which reflects the exact scoring used at each search step. This also reveals two  reasons from decoding perspective why applying ILM can improve the recognition performance: 
\vspace{-1.2mm}
\begin{enumerate}[itemsep=-0.5mm]
\item [R1.] The label distribution of $P_{\text{RNNT}}$ is rebalanced with the prior removal, so that we rely more on the external LM for context modeling, which is a desired behavior.
\item [R2.] The division by $P_{\text{ILM}}$ boosts the label probability against the (usually high) blank probability, so that the importance of the external LM $({\lambda}1)$ can be increased without suffering a large increment of deletion errors.
\end{enumerate}
\vspace{-1.2mm}
The R2 explains why SF (${\lambda}2 = 0$) can only achieve a limited performance. It may also alleviate the necessity of heuristic approaches in decoding such as length-normalization \cite{graves2012rnnt, mWER_HAT} and length-reward \cite{CtcLenReward, McDermott2019DR, saon21rnnt}.
However, an increasing ${\lambda}2$ with more boosting can also lead to an increment of insertion and/or substitution errors. Therefore, both scales require careful tuning in practice.

\subsection{Exact-ILM Training}
In the appendix A of \cite{variani2020hat}, a detailed proof is given to show:\\
\scalebox{0.9}{\parbox{1.11\linewidth}{%
\begin{align*}
P_{\text{RNNT-ILM}}(a_s | a_1^{s-1}) \propto \exp \left[ J_{\backslash\epsilon}\left( f^{\text{pred}}(a_1^{s-1}) \right) \right] \numberthis \label{eq:proof}
\end{align*}}}
if the following assumption can hold:\\
\scalebox{0.9}{\parbox{1.11\linewidth}{%
\begin{align*}
J_{\backslash\epsilon}\left( f^{\text{pred}}(a_1^{s-1}), f_t^{\text{enc}}(X) \right) = J_{\backslash\epsilon}\left( f^{\text{pred}}(a_1^{s-1}) \right) + J_{\backslash\epsilon}\left(f_t^{\text{enc}}(X) \right) 
\end{align*}}}
\Cref{eq:proof} then leads to the $h'_{\text{zero}}$ ILM approach.
However, this assumption does not always hold if $J$ contains some non-linearity, which is mostly the case for better performance. 

Here we propose to further generalize the assumption:\\
\scalebox{0.9}{\parbox{1.11\linewidth}{%
\begin{align*}
J_{\backslash\epsilon}\left( f^{\text{pred}}(a_1^{s-1}), f_t^{\text{enc}}(X) \right) = J'(a_1^{s-1}) + J_{\backslash\epsilon}\left(f_t^{\text{enc}}(X) \right) \numberthis \label{eq:assump2}
\end{align*}}}
where $J'$ can be any function with output size $|V|$.
As long as $J'$ is independent of $X$, the proof still holds and we have:\\
\scalebox{0.9}{\parbox{1.11\linewidth}{%
\begin{align*}
P_{\text{RNNT-ILM}}(a_s | a_1^{s-1}) \propto \exp \left[ J'(a_1^{s-1}) \right] \numberthis \label{eq:proof2}
\end{align*}}}
Instead of relying on the assumption, we can train $J'$ to fulfill \Cref{eq:assump2} towards an exact ILM estimation eventually. 
Besides the mean squared error loss, this exact-ILM training objective $\L_{J'}$ can also be the cross entropy (CE) loss, since both sides of \Cref{eq:assump2} are actually the logits output over $V$, which we can directly apply softmax. 
The CE loss may introduce a constant shift to \Cref{eq:assump2}, which is still valid for the proof in HAT \cite{variani2020hat} to achieve \Cref{eq:proof2}.
Additionally, by performing Viterbi alignment of the acoustic data with a (pre)trained RNN-T model, $\L_{J'}$ can be further simplified to consider only those $h_t$ where $a_s$ occurs. 

Among various possibilities, we can directly apply this exact-ILM training to the $h'_{a_1^{s-1}}$ approach by defining:\\
\scalebox{0.9}{\parbox{1.11\linewidth}{%
\begin{align*}
J'(a_1^{s-1}) = J_{\backslash\epsilon}\left( f^{\text{pred}}(a_1^{s-1}), f_{\theta_{\text{ILM}}}(a_1^{s-1}) \right)
\end{align*}}}
and train $f_{\theta_{\text{ILM}}}$ by the following multi-task training:\\
\scalebox{0.9}{\parbox{1.11\linewidth}{%
\begin{align*}
\L^{\text{exact}}_{\text{ILM}} = \L_{\text{ILM}} + \alpha \L_{J'}
\end{align*}}}
This extension directly enables a theoretical justification for the $h'_{a_1^{s-1}}$  ILM approach, which is trained towards \Cref{eq:proof2}. 
Similar as $\L_{\text{ILMT}}$, we can also combine $\L_{\text{RNNT}}$ and $\L_{J'}$ to have a joint training of all parameters, where the model is additionally forced to better fulfill the assumption towards an exact ILM estimation. This may require a careful design of $J'$, which is not investigated in this work.


\section{Experiments}
\subsection{Setup}
Experimental verification is done on the 960h Librispeech corpus \cite{libsp} for in-domain evaluation, and on the TLv2 corpus \cite{tedlium2} for out-of-domain evaluation. We use 5k acoustic data-driven subword modeling (ADSM) units \cite{zhou2021ADSM} obtained from the Librispeech corpus. 
We follow the ADSM text segmentation and train two individual external LMs on the corresponding LM data of each corpus.
The Librispeech LM contains 32 Transformer layers \cite{irie19trafolm} and the TLv2 LM contains 4 long short-term memory (LSTM) \cite{hochreiter1997lstm} layers with 2048 units.

We use 50-dim gammatone features \cite{schluter2007gt} and a strictly monotonic ($U=T$) version of RNN-T \cite{tripathi2019monoRNNT}.
The encoder $f^{\text{enc}}$ contains 2 convolutional layers followed by 6 bidirectional-LSTM (BLSTM) layers with 640 units for each direction. A subsampling of factor 4 is applied via 2 max-pooling layers in the middle of BLSTM stacks. The prediction network $f^{\text{pred}}$ contains an embedding layer of size 256 and 2  LSTM layers with 640 units. We use the standard additive joint network for $J$ which contains 1 linear layer of size 1024 followed by the tanh activation, and another linear layer followed by the final softmax. The RNN-T model is only trained on the Librispeech corpus. We firstly apply a Viterbi training variant \cite{zhou2021phonemeTransducer} for 30 full epochs and then fine-tune the model with $\L_{\text{RNNT}}$ for 15 full epochs. This model is used as the base model for all further experiments.

The density ratio \cite{McDermott2019DR} LM uses the same structure as $f^{\text{pred}}$.
For the $h'_{a_1^{s-1}}$ approach, we apply the mini-LSTM network \cite{zeineldeen2021ilm} for $f_{\theta_{\text{ILM}}}$ which is trained with either $\L_{\text{ILM}}$ or $\L^{\text{exact}}_{\text{ILM}}$ for 0.5-1 full epoch on Librispeech. 
For the $\L^{\text{exact}}_{\text{ILM}}$, we use the CE loss for $\L_{J'}$, and $\alpha=1.0$ and $2.0$ for in-domain and cross-domain evaluation, respectively.
The base RNN-T model is used to generate the Viterbi alignment for $\L_{J'}$ simplification.

For $\L_{\text{ILMT}}$, we follow \cite{Meng2021ILMT} to initialize with the base model and adjust $\alpha=0.2$.
To avoid the potential improvement just form a much longer re-training with learning rate reset \cite{vieting2021hybrid}, we only apply fine-tuning with $\L_{\text{ILMT}}$ for upto 10 more full epochs on Librispeech. 
Since $\L_{\text{ILM}}$ is only relevant for $f^{\text{pred}}$ and $J$, we also freeze $f^{\text{enc}}$ during this procedure. 
All the best epochs based on no-LM recognition show the same or slightly better performance than the base model, which are used for further LM integration evaluation.

The decoding follows the description in \Cref{sec:decode}.
We apply alignment-synchronous search \cite{saon2020ALSD} with score-based pruning and a beam limit of 128.
We explicitly do not apply any heuristic approach for decoding to better reflect the effect of each LM integration method.
All scales are optimized on the dev sets using grid search.

\begin{table}[t!]
\caption{\it WER[\%] results of LM integration evaluation on the in-domain Librispeech and out-of-domain TLv2 corpora.}
\vspace{-1mm}
\setlength{\tabcolsep}{0.25em}
\begin{center}\label{tab:WER}
\begin{tabular}{|l|c|c|c|c|c||c|c|}
\hline
\multirow{3}{*}{\shortstack[c]{\hspace{1mm}Model\\\hspace{1mm}Train}} & \multirow{3}{*}{Evaluation} & \multicolumn{4}{c||}{Librispeech} & \multicolumn{2}{c|}{TLv2} \\ \cline{3-8}
& & \multicolumn{2}{c|}{dev} & \multicolumn{2}{c||}{test} & \multirow{2}{*}{dev} & \multirow{2}{*}{test} \\
& & clean & other & clean & other & & \\ \hline
\multirow{5}{*}{$\L_{\text{RNNT}}$} & no LM & 3.3 & 9.7 & 3.6 & 9.5 & 19.8 & 20.3 \\ \cline{2-8}
 & SF & 2.0 & 5.1 & 2.2 & 5.5 & 15.5 & 16.4 \\ \cline{2-8}
 & density ratio & 1.9 & 4.8 & 2.1 & 5.2 & 14.1 & 15.0 \\ \cline{2-8}
 & $h'_{\text{zero}}$ & 1.8 & 4.4 & 2.0 & 4.8 & 13.6 & 14.4 \\ \cline{2-8}
 & $h'_{\text{avg}}$ & 1.8 & 4.4 & 2.0 & 4.9 & 13.5 & 14.6 \\ \hline
\hspace{1mm}+ $\L_{\text{ILM}}$ & \multirow{2}{*}{$h'_{\text{mini-LSTM}}$} & 1.8 & 4.3 & 1.9 & 4.7 & 13.4 & 14.4 \\ \cline{1-1} \cline{3-8}
\hspace{1mm}+ $\L^{\text{exact}}_{\text{ILM}}$ & & \bf1.8 & \bf4.2 & \bf1.9 & \bf4.6 & \bf13.2 & \bf14.0 \\ \hline
\hline
\multirow{3}{*}{$\L_{\text{ILMT}}$} & $h'_{\text{zero}}$ & 1.8 & 4.4 & 2.0 & 4.8 & 13.3 & 14.2 \\ \cline{2-8}
 & $h'_{\text{avg}}$ & 1.9 & 4.5 & 2.1 & 4.9 & 13.5 & 14.4 \\ \cline{2-8}
 & $h'_{\text{mini-LSTM}}$ & 1.8 & 4.4 & 2.0 & 4.8 & 13.2 & 14.1 \\ \hline
\end{tabular}
\end{center}
\vspace{-4mm}
\end{table}

\subsection{LM Integration Evaluation} 
\Cref{tab:WER} shows the word error rate (WER) results of the aforementioned LM integration methods evaluated on the in-domain Librispeech and out-of-domain TLv2 tasks.
As expected, the external LMs bring significant improvement over the standalone RNN-T, and all ILM correction approaches improves further over the simple SF.
For both tasks, all $h'$-based methods outperform the density ratio approach. 
Both $h'_{\text{zero}}$ and $h'_{\text{avg}}$ show similar performance, and the $h'_{\text{mini-LSTM}}$ trained with $\L_{\text{ILM}}$ performs slightly better than the two.
The proposed $\L^{\text{exact}}_{\text{ILM}}$ further improves the $h'_{\text{mini-LSTM}}$ approach.

The $\L_{\text{ILMT}}$ improves all three $h'$-based approaches on the cross-domain TLv2 task, while no positive effect is obtained on the in-domain Librispeech task.
This is in line with the observation in HAT \cite{variani2020hat}, where a decreasing $\L_{\text{ILM}}$ leads to no improvement on the overall performance. 
T$  $he $\L^{\text{exact}}_{\text{ILM}}$ trained $h'_{\text{mini-LSTM}}$ approach performs the best in all cases.

\subsection{Analysis}
To verify the 2 decoding-perspective benefits of ILM correction as claimed in \Cref{sec:decode}, we conduct additional analytical experiments on the Librispeech dev-other set using the base RNN-T model. 
To simulate the effect of boosting label probability (R2) without the effect of rebalanced label distribution (R1), we apply a constant length reward upon SF for decoding. To simulate the effect of R1 without the effect of R2, we apply a modified $h'_{\text{zero}}$ evaluation as following. For each $y_u \neq \epsilon$, we firstly apply a renormalization as:\\
\scalebox{0.9}{\parbox{1.1\linewidth}{%
\begin{align*}
P_{\text{norm}}(y_u) = \frac{P_{\text{RNNT}}(y_u | \B(y_1^{u-1}), h_1^T) / P^{{\lambda}2}_{\text{ILM}}(y_u | \B(y_1^{u-1}))}{\sum_{a \in V} P_{\text{RNNT}}(a | \B(y_1^{u-1}), h_1^T) / P^{{\lambda}2}_{\text{ILM}}(a | \B(y_1^{u-1}))}
\end{align*}}}
Then we modify the probability of $y_u \neq \epsilon$ used for search as:
\scalebox{0.9}{\parbox{1.1\linewidth}{%
\begin{align*}
\left( 1 - P_{\text{RNNT}}(\epsilon | \B(y_1^{u-1}), h_1^T) \right) \cdot P_{\text{norm}}(y_u) \cdot P^{{\lambda}1}_{\text{LM}}(y_u | \B(y_1^{u-1}))
\end{align*}}}
The first two terms of the product effectively restrict the label probability back to a single distribution w.r.t. the blank probability of $P_{\text{RNNT}}$, but still maintain the rebalanced label distribution to some extent. We denote this operation as $\text{renorm-}{\epsilon}$. 
\Cref{tab:analysis} shows the WER, substitution (Sub), deletion (Del) and insertion (Ins) error rate results together with the optimized scales for these experiments.

For the baseline SF, the optimal ${\lambda}1$ already leads to a high Del error and we can not increase it further. Boosting the label probability with length reward largely reduces the Del error to the same level as Ins. It also allows a slight increase of the external LM importance (${\lambda}1$) for better performance. This verifies the individual effect of R2. 
Rebalancing the label distribution with $h'_{\text{zero}} + \text{renorm-}{\epsilon}$ reduces the Sub error as we rely more on the external LM for context modeling. However, it still suffers the high Del error without the boosting effect. This verifies the individual effect of R1.
When combining length reward and $h'_{\text{zero}} + \text{renorm-}{\epsilon}$, we see that the benefits are complementary.
Finally, applying the $h'_{\text{zero}}$ ILM correction allows further enlarging the effect of R1 and R2 with larger scales, and thus achieves further improvement. It also eliminates the need of length reward.

\begin{table}[t!]
\caption{\it WER[\%] and Sub,Del,Ins error rate[\%] results on the dev-other set of the in-domain Librispeech corpus. Analytical evaluation using the base RNN-T model.}
\vspace{-1mm}
\setlength{\tabcolsep}{0.5em}
\begin{center}\label{tab:analysis}
\begin{tabular}{|l|c|c|c||c|c|c|}
\hline
\multirow{2}{*}{Evaluation} & \multirow{2}{*}{${{\lambda}1}$} & \multirow{2}{*}{${{\lambda}2}$} & \multicolumn{4}{c|}{Librispeech dev-other} \\ \cline{4-7}
 & & & WER & Sub & Del & Ins \\ \hline
SF & 0.61 & \multirow{2}{*}{0} & 5.1 & 3.8 & 0.9 & 0.4 \\ \cline{1-2} \cline{4-7}
\hspace{1mm} + length reward & 0.65 & & 4.8 & 3.8 & 0.5 & 0.5 \\ \hline \hline
$h'_{\text{zero}} + \text{renorm-}{\epsilon}$ & 0.61 & \multirow{2}{*}{0.35} & 4.9 & 3.6 & 0.9 & 0.4 \\ \cline{1-2} \cline{4-7}
\hspace{1mm} + length reward & 0.65 &  & 4.6 & 3.7 & 0.5 & 0.4 \\ \hline \hline
$h'_{\text{zero}}$ & 0.85 & \multirow{2}{*}{0.4} & \bf4.4 & \bf3.5 & \bf0.5 & \bf0.4 \\ \cline{1-2} \cline{4-7}
\hspace{1mm} + length reward & 0.95 &  & 4.5 & 3.6 & 0.5 & 0.4  \\ \hline
\end{tabular}
\end{center}
\vspace{-5.5mm}
\end{table}

\vspace{-1mm}
\section{Conclusion}
\vspace{-0.5mm}
In this work, we provided a detailed formulation to compare various ILM correction-based LM integration methods in a common RNN-T framework. We explained two major reasons for performance improvement with ILM correction from a decoding interpretation, which are experimentally verified with detailed analysis. Moreover, we proposed an exact-ILM training framework by extending the proof in HAT \cite{variani2020hat}, which enables a theoretical justification for other ILM approaches.
All investigated LM integration methods are systematically compared on the in-domain Librispeech and out-of-domain TLv2 tasks.
The recently proposed $h'_{\text{mini-LSTM}}$ ILM approach for the attention model also performs the best for the RNN-T model. Our proposed exact-ILM training can further improve its performance.

\begin{center}
\bf{\normalsize{Acknowledgements}}
\end{center}
\vspace{-0.6mm}
\footnotesize
This work was partly funded by the Google Faculty Research Award for ``Label Context Modeling in Automatic Speech Recognition".
We thank Mohammad Zeineldeen and Wilfried Michel for useful discussion.

\let\normalsize\small\normalsize
\let\OLDthebibliography\thebibliography
\renewcommand\thebibliography[1]{
        \OLDthebibliography{#1}
        \setlength{\parskip}{-0.3pt}
        \setlength{\itemsep}{1pt plus 0.07ex}
}

\bibliographystyle{IEEEbib}
\bibliography{refs}

\end{document}